%% file: example_paper.tex
\documentclass{article}

\usepackage{microtype}
\usepackage{graphicx}
\usepackage{subfigure}
\usepackage{booktabs} 
\usepackage{bm}
\usepackage{amsfonts,amssymb}
\usepackage{hyperref}

\newcommand{\ourname}[0]{AutoDistill}
\newcommand{\tpu}[0]{TPUv4i}

\usepackage[accepted]{mlsys2022}

\mlsystitlerunning{\ourname: an End-to-End Framework to Explore and Distill Hardware-Efficient Language Models}

\begin{document}

\twocolumn[
\mlsystitle{\ourname: an End-to-End Framework to Explore and Distill Hardware-Efficient Language Models}




\begin{mlsysauthorlist}
\mlsysauthor{Xiaofan Zhang$^1$, Zongwei Zhou$^2$, Deming Chen$^1$, Yu Emma Wang$^2$}{}\\ \vspace{+0.5em}
\mlsysauthor{\rm{$^1$University of Illinois Urbana-Champaign} \rm\textit{\{xiaofan3, dchen\}@illinois.edu}}{}\\
\mlsysauthor{\rm{$^2$Google} \rm\textit{\{zongweiz, yuemmawang\}@google.com}}{}
\end{mlsysauthorlist}



\mlsyskeywords{Machine Learning, MLSys}

\vspace{+8pt}
\begin{abstract}
\vspace{+6pt}
Recently, large pre-trained models have significantly improved the performance of various Natural Language Processing (NLP) tasks but they are expensive to serve 
due to long serving latency and large memory usage. 
To compress these models, knowledge distillation has attracted an increasing amount of interest as one of the most effective methods for model compression.
However, existing distillation methods have not yet addressed the unique challenges of model serving in datacenters, such as handling fast evolving models, considering serving performance, and optimizing for multiple objectives.
To solve these problems,
we propose \ourname, an end-to-end model distillation framework integrating model architecture exploration and multi-objective optimization for building hardware-efficient NLP pre-trained models. We use Bayesian Optimization (BO) to conduct multi-objective Neural Architecture Search (NAS) for selecting student model architectures. The proposed search comprehensively considers both prediction accuracy and serving latency on target hardware. We propose Flash Distillation, a model-agnostic technique using a much shorter period of progressive knowledge transfer to distinguish promising student model candidates from less promising ones. Together with the BO algorithm, it significantly reduces the cost during model exploration. 
The experiments on {\tpu}~\cite{jouppi2021ten} show the finding of seven model architectures with better pre-trained accuracy (up to 3.2\% higher) and lower inference latency (up to 1.44$\times$ faster) than MobileBERT \cite{sun2020mobilebert}. 
By running nine downstream NLP tasks in the GLUE benchmark, the model distilled for pre-training by {\ourname} with 28.5M parameters achieves an 81.69 average score, which is higher than BERT$\rm_{BASE}$~\cite{devlin2018bert}, DistillBERT~\cite{sanh2019distilbert}, TinyBERT~\cite{jiao2020tinybert}, NAS-BERT~\cite{xu2021nasbert}, and MobileBERT~\cite{sun2020mobilebert}. 
The most compact model found by {\ourname} contains only 20.6M parameters but still outperform BERT$\rm_{BASE}$ (109M), DistillBERT (67M), TinyBERT (67M), and MobileBERT (25.3M) regarding the average GLUE score.
By evaluating on SQuAD, a model found by {\ourname} achieves an 88.4\% F1 score with 22.8M parameters, which reduces parameters by more than 62\% while maintains higher accuracy than 
DistillBERT, 
TinyBERT, 
and NAS-BERT.
\end{abstract}
]




\input{sec_introduction}
\input{sec_relatedwork}
\input{sec_proposeddesign}

\input{sec_experiment}
\input{sec_conclusion}
\input{sec_acknowledge}


\bibliography{example_paper}
\bibliographystyle{mlsys2022}



\end{document}

%% file: sec_introduction.tex
\section{Introduction}
\label{sec:intro}

In recent years, large-scale pre-trained language models have achieved state-of-the-art results on many tasks. These models not only facilitate a variety of Natural Language Processing (NLP) applications but also have continuously improved the result quality of these challenging tasks~\cite{vaswani2017attention,peters2018deep,yang2019xlnet,devlin-etal-2019-bert}. Among these models, BERT~\cite{devlin-etal-2019-bert} achieves state-of-the-art performance on a number of NLP tasks and has profoundly affected subsequent model designs~\cite{liu2019roberta, zhang2019ernie, Lan2020albert,sun2020mobilebert}.

With the advent of such large-scale language models, minimizing the serving cost is becoming increasingly important.
Traditionally, serving optimization has been studied in the context of mobile or embedded devices where available resources are heavily constrained.
On the other hand, large-scale models make serving challenging even for datacenters due to their sheer sizes.
For example, GPT-3, a BERT-like model with 175 billion weights~\cite{floridi2020gpt},
is over 500$\times$ larger than BERT~\cite{devlin2018bert}, 
and it likely needs 500$\times$ compute and memory resources.
Furthermore, recent advances in techniques to amortize the training cost of large language models such as fine-tuning make serving even more costly than training, let alone the fact that training cost is usually amortized over weeks or months while the cost of serving adds up from every request.

To alleviate this problem, recent work has extensively investigated model compression techniques, such as parameter quantization~\cite{shen2020qbert}, network pruning~\cite{gordon2020compressing}, and knowledge distillation~\cite{sanh2019distilbert,sun2020mobilebert,jiao2020tinybert,chen2020adabert}.
Among them, knowledge distillation is a promising technique that compresses large models by generating a compact model (the student model) and training it based on the trained pattern of a larger model (the teacher model)~\cite{hinton2015distilling}.
Earlier work has focused on distilling large models, like BERT, to task-specific compact designs with less redundancy in model architecture~\cite{sun2019patient,tang2019distilling, tsai2019small}, or task-agnostic pre-trained models, which can then be fine-tuned to different downstream tasks~\cite{sun2020mobilebert,xu2021nasbert}. 



In this paper, we address two key challenges in practical application of knowledge distillation as part of regular large-scale model release processes: (1) fully automated, efficient distillation process and (2) latency-guided model optimization. Production models in datacenters are diverse and evolve rapidly, which necessitates distillation without human in the loop for scalability. Also, latency is a crucial metric for production model serving since user-facing products often have strict latency requirements and any latency reduction leads to a significant cost reduction considering datacenters' large volume.

Unfortunately, existing knowledge distillation approaches do not meet the aforementioned requirements, making it not suitable for production deployment.
First, prior work requires manual effort to design a tailored student model for a given teacher model architecture, which limits its scalability and time to market.
Manually designed student models~\cite{sanh2019distilbert,sun2019patient,jiao2020tinybert,sun2020mobilebert}, chosen by domain experts with a few iterations of trial and error, can result in sub-optimal performance with large sizes and low accuracy and may not adapt well to new teacher model architectures.
NAS~\cite{chen2020adabert,xu2021nasbert,gao2021autobert} helps explore larger model architecture design space, but the search algorithms used by existing efforts need to be crafted for target models or objectives.
More importantly, the challenging supernet training on NLP pre-training tasks makes it difficult to scale to larger architecture design spaces.

Second, existing distillation work is not aware of latency and mainly formulates model compression as a single objective optimization problem that minimizes the model size with accuracy loss as a constraint~\cite{sanh2019distilbert,sun2019patient,jiao2020tinybert}.
This leads to suboptimal serving performance since reduced model sizes do not necessarily lead to lower serving latency.
While some previous work uses proxies~\cite{tsai2020finding, xu2021nasbert} to approximate hardware performance, e.g., weighted sum of the operation latency, number of multiply-accumulate operations, and number of model parameters, such analytical models are often not accurate because the serving latency of models can vary significantly based on the hardware and software stack.

To this end, we propose \ourname, a model distillation framework integrating model architecture exploration and multi-objective optimization for building hardware-aware NLP pre-trained models.
To summarize, the main contributions of this paper are as follows
\begin{enumerate}
    \setlength{\itemsep}{-2pt}
    \vspace{-6pt}
    \item We propose an end-to-end framework for fully automated model distillation, which satisfies user-defined metrics and constraints by delivering optimized pre-trained models distilled from large NLP models.
    It can be easily extended to new search spaces and objectives, thereby eliminating the need for distillation experts. It helps solve the most critical problem of productionizing large-scale model distillation in datacenters.
    \item We use Bayesian Optimization (BO)~\cite{golovin2017vizier,snoek2012practical} to conduct multi-objective NAS for student model architectures. The proposed search comprehensively considers both prediction accuracy and serving latency on target serving hardware.
    It is the first time that BO is adopted by the NAS and distillation framework to deliver hardware-efficient large-scale NLP pre-trained models. 
    %
    \item Enabled by \ourname, the experiments on {\tpu} 
    identify
    seven model architectures with up to 3.2\% higher pre-trained accuracy and up to 1.44$\times$ speedup on latency compared to MobileBERT~\cite{sun2020mobilebert}.
    Four of them have higher GLUE average scores (up to 81.69) than BERT$\rm_{BASE}$~\cite{devlin2018bert}, DistillBERT~\cite{sanh2019distilbert}, TinyBERT~\cite{jiao2020tinybert}, and MobileBERT. 
    Two models are smaller and have higher SQuAD accuracy than DistillBERT, TinyBERT, and NAS-BERT~\cite{xu2021nasbert}. 
\end{enumerate}


%% file: sec_relatedwork.tex
\section{Background and Related Work}
\label{sec:relatedwork}

After BERT was proposed in \cite{devlin-etal-2019-bert}, it has attracted extensive studies on model compression. Knowledge distillation is one widely adopted method to deliver compact BERT models for serving environments where memory or latency is limited. 
For example, \citet{tang2019distilling} perform task-specific knowledge distillation and transfer the knowledge from BERT to a single-layer LSTM model while \citet{tsai2019small} distill smaller BERT models for sequence labeling tasks. 
\citet{sun2019patient} develop a distillation method to extract knowledge from a teacher model's intermediate and last layers.
DistillBERT \cite{sanh2019distilbert} performs distillation during model pre-training and reduces the depth of BERT by half. TinyBERT \cite{jiao2020tinybert} performs layer-wise distillation for model pre-training and fine-tuning for more aggressive compression. To better adopt different tasks, \citet{hou2020dynabert} propose adaptive compressed models for specific downstream tasks.
Researchers also focus on building task-agnostic compressed models. For example, \citet{sun2020mobilebert} propose MobileBERT that can be generically fine-tuned on different downstream NLP tasks. It adopts a bottleneck structure to reduce a model's layer dimension for compressed overall model size.    
However, these
manually designed student models can lead to sub-optimal performance, and causes high barrier and overhead for distilling new models.

Recent work has shown increasing 
interest in leveraging
NAS for NLP model compression with the goal to discover more diverse model architectures, so that models no longer rely on handcrafted designs. For example, 
\citet{tsai2020finding} use a one-shot architecture search to find Transformer-like model structures for specific NLP tasks. 
\citet{chen2020adabert} adopt NAS to search for task-specific model architectures and incorporate knowledge distillation to let student models capture the task-oriented knowledge from their teacher. 
As a comparison, we target delivering task-agnostic models. They are more challenging to distill while achieving the same prediction accuracy than the task-specific ones, but they can be generally applicable to different downstream tasks via simple fine-tuning, which is more desirable for users to simplify the fine-tuning pipeline.

More recently, NAS-BERT adopts NAS for compressed task-agnostic models \cite{xu2021nasbert}.
Their NAS algorithms are not as flexible as black-box optimization algorithms such as BO 
since their design space can not be easily scaled up. The main problem is that the supernet (which contains all possible operations in the design space) is too large to train on NLP pre-training tasks and requires customized compression by reducing the design space. By contrast, it is easier for BO to scale up to a larger design space and more objectives as supernet training is not required.

Although knowledge distillation and NAS help diversify the compressed model architectures, existing work mainly focuses on reducing model sizes with accuracy loss as a constraint and formulates model compression as a single-objective optimization problem. 
Such a strategy may not guarantee that the compressed model can be efficiently deployed on the target hardware as smaller models do not necessarily perform faster. 
To address this problem, \citet{li2021searching} create a datacenter-optimized network search space and adopt NAS to discover neural networks with optimized accuracy and serving latency. 
NAS-BERT \cite{xu2021nasbert} adopts a look-up table (LUT) to calculate the overall inference latency by adding up costs of the selected operations and uses the latency to guide model search. Similarly, \citet{tsai2020finding} consider the layer-wise hardware costs and calculate the overall model performance as a weighted summation of these costs during architecture search. However, the weighted summation represents the cost expectation of a group of operations rather than the actual cost on the selected architecture. It also excludes the memory transfer overhead between operations which is very important for accurately modeling performance.

In this paper, we intend to adopt precise 
hardware feedback by using models' measured 
hardware performance while running on the target hardware to guide the multi-objective architecture search. 
Compared to previous designs using proxy or approximated hardware feedback, our method captures more hardware information that cannot be obtained by previous methods. 
To reduce the search cost, we propose Flash Distillation and adopt BO algorithm to effectively explore model architectures.  
All these features are integrated in our proposed framework, {\ourname}, to deliver task-agnostic hardware-efficient pre-trained models.

%% file: sec_proposeddesign.tex
\section{\ourname}
\label{sec:proposeddesign}


\subsection{Framework Overview}

\begin{figure*}
    \centering
    \vspace{-6pt}
    \includegraphics[width=0.85\textwidth]{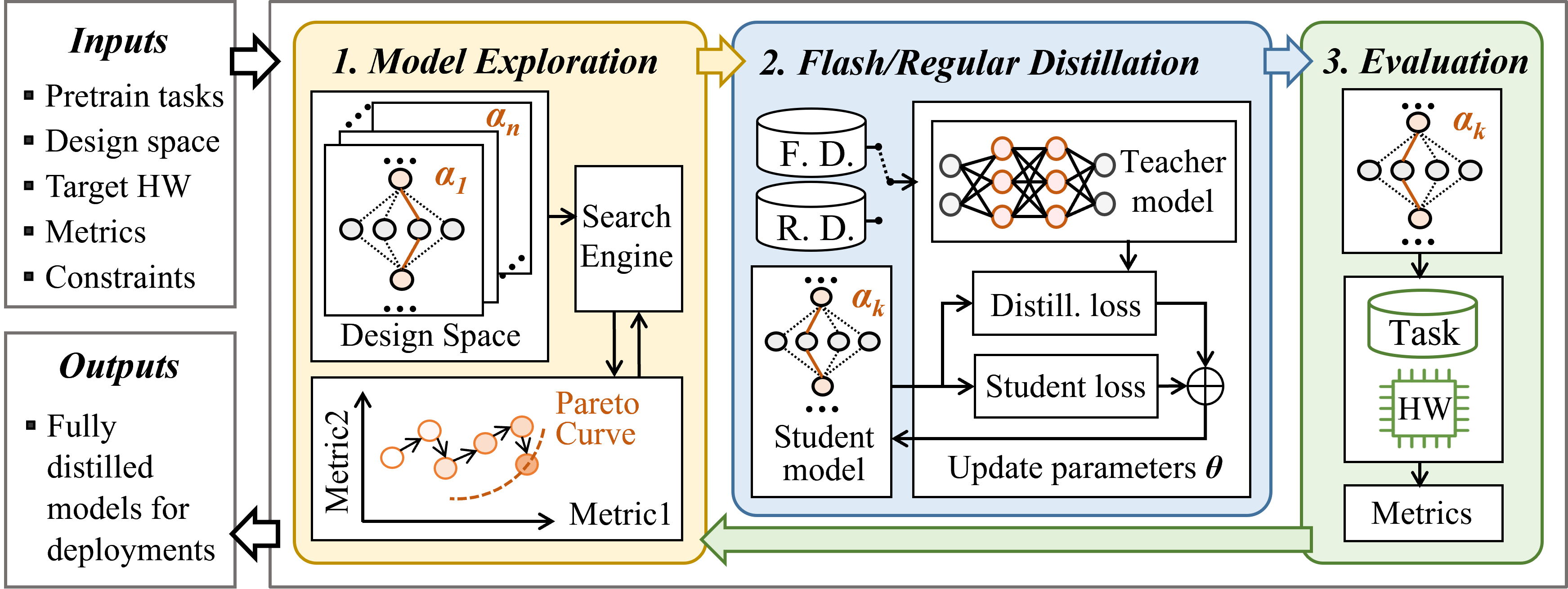}
    \vspace{-2pt}
    \caption{Illustration of the proposed {\ourname} framework. \textit{Model Exploration}, \textit{Flash Distillation}, and \textit{Evaluation} are applied iteratively for compressed model exploration. After exploration, the selected models are passed to regular distillation and then output by the framework. We assume $n$ different architectures are defined in the design space and model $a_k$ is proposed in the $k$-th iteration. F. D. and R. D. represent the training setups for Flash Distillation and regular distillation.}
    \vspace{-6pt}
    \label{fig:overall_flow}
\end{figure*}

{\ourname} provides an end-to-end solution to satisfy user requirements and generates optimized task-agnostic pre-trained models for target 
hardware. User requirements, including objectives and constraints, are passed to {\ourname} as inputs, which include pre-training tasks, model design spaces, target hardware, evaluation metrics, and constraints (e.g., model size, inference latency limit) that need to be considered.  
We illustrate the overall flow in Figure \ref{fig:overall_flow}.
Three major stages, as \textit{Model Exploration}, \textit{Flash Distillation}, and \textit{Evaluation}, are executed in a loop for searching models that best suit the user inputs. Every time the loop is completed, we call it one iteration.
After several iterations, the search engine returns the models on the Pareto curve and they are passed to \textit{Regular Distillation} for more thorough pre-training, so that they can be prepared to serve different downstream tasks.

In \textit{Model Exploration}, the architecture design space is first initialized and passed to the search engine. During every iteration, the engine searches for a better compressed model by considering the design space, evaluation metrics, and user-specified constraints. We list two metrics as examples in Figure \ref{fig:overall_flow} (for Metric 1, the higher the better; while for Metric 2, the lower the better). After several iterations, found models are plotted on the same coordinate, and {\ourname} selects those located along the Pareto curve as the most promising candidates. We will provide more detailed explanations regarding the design space and the search algorithm design in Subsection \ref{sec:design_space} and \ref{sec:nas_algorithm}, respectively.

{\ourname} then adopts \textit{Flash Distillation} to grow the model recommended by the last stage. 
This model is considered as a student model, which learns from both pre-training datasets and the teacher model.
We include three knowledge distillation technologies: a layer-wise knowledge transfer,
a progressive knowledge transfer, 
and a model pre-training distillation
(details in Subsection \ref{sec:flash_disllation}). 
We have demonstrated that the Flash Distillation needs only 5\% of the regular pre-training steps to distinguish promising models at the early stage, which significantly reduces the search efforts.
This stage is also responsible for Regular Distillation with the same teacher model but different training setups (which are the hyperparameters, e.g., training steps, learning rate, and batch size). After iterations of model exploration, Regular Distillation is launched with more thorough 
pre-training setups (e.g., more training steps) than Flash Distillation.
After that, {\ourname} output fully distilled models.

In \textit{Evaluation}, the flash-distilled student model is evaluated with the target tasks and hardware. In general, commonly used metrics include the prediction accuracy (e.g., masked language modeling (MLM) accuracy, next sentence prediction (NSP) accuracy) and the hardware performance (e.g., inference latency, FLOPs utilization, maximum memory footprint). After all desired metrics are collected, all information is passed to the \textit{Model Exploration} stage and the search engine selects the next model for the next iteration.     
We will introduce how we capture the precise hardware performance in Subsection \ref{sec:hw_feedback}.

\subsection{Student model template and model architecture design space}
\label{sec:design_space}

\begin{figure}[t]
    \centering
    \includegraphics[width=0.43\textwidth]{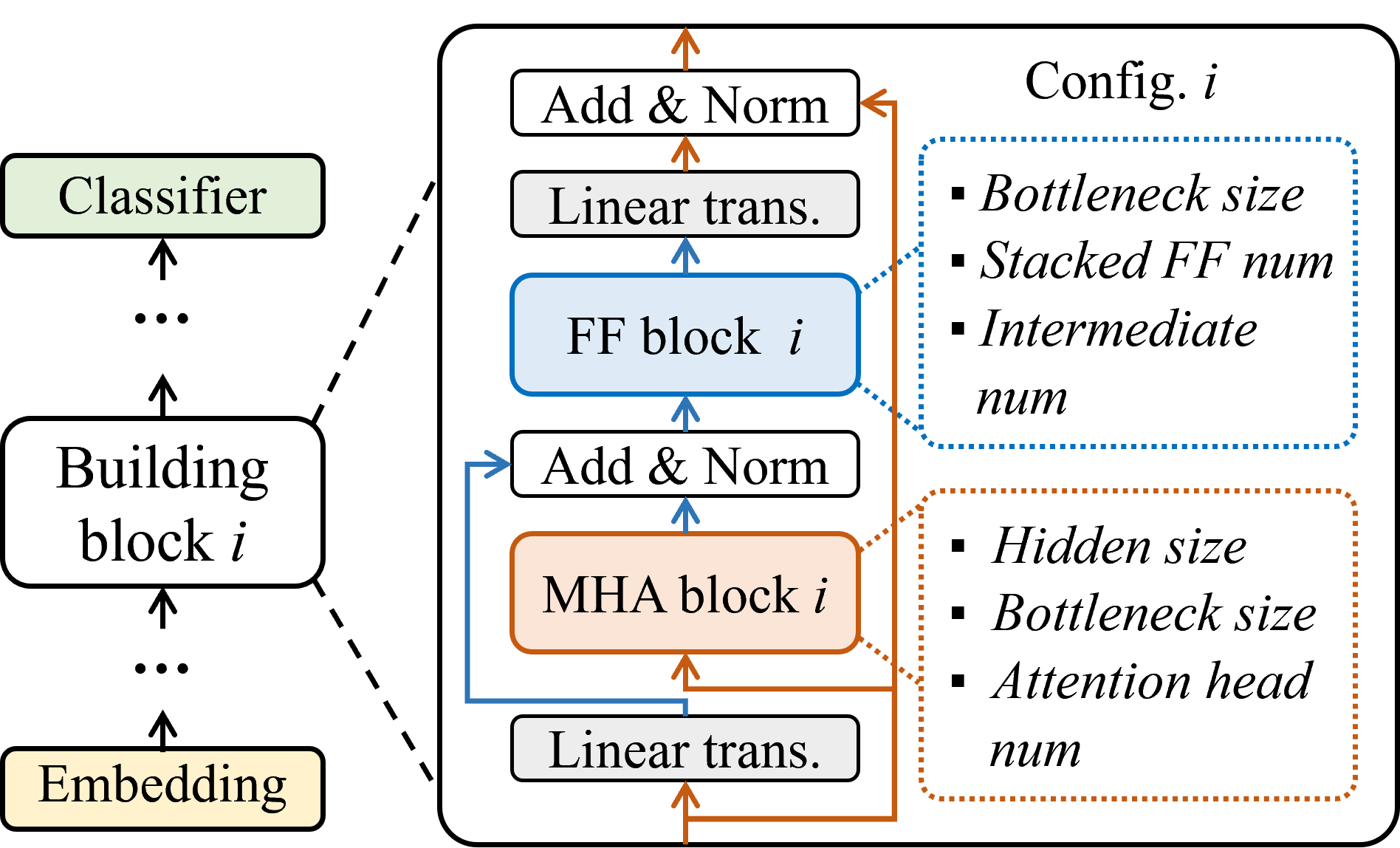}
    \caption{The student model design template adopted by {\ourname} for model exploration. It contains a bottleneck structure between two linear transformations and two major configurable blocks including the multi-head attention (HMA) block and the feed-forward (FF) block~\cite{sun2020mobilebert}.
    The orange and blue arrows represent data flows with different dimensions specified by \textit{Hidden Size} and \textit{Bottleneck Size}.
    }
    \vspace{-8pt}
    \label{fig:model_template}
\end{figure}

To enable more diverse model configurations, we refer to the bottleneck structure from \cite{sun2020mobilebert} and build a flexible student model template as shown in Figure \ref{fig:model_template}. 
It defines the student model design space by providing multiple configurable network components. With the proposed template, {\ourname} can search for the most suitable model to satisfy user-specific requirements. 
The proposed template follows a chain-structure and consists of a stack of $n$ configurable building blocks, which connect the first embedding layer and the last classifier. It shares a similar block-based network design as BERT to ensure effective knowledge transfer from BERT-like teacher models. 
For each building block, taking the building block $i$ as an example, there are two major configurable blocks, called multi-head attention (MHA) and feed-forward (FF). These two blocks contribute to a five-dimensional architecture design space which provides a variety of structure combinations.

We summarize the proposed design space in Table \ref{tab:design_space}. The first configurable factor is called \textit{Hidden Size}, which indicates the input and output dimensions of the building block and the input dimension of the MHA block. The second factor is \textit{Bottleneck Size}, which shows the output dimension of the MHA block. 
We have drawn orange and blue lines in Figure \ref{fig:model_template} to respectively denote data flows related to the configuration of hidden size and bottleneck size. By taking advantage of two linear transformation layers, the proposed student model design template can accommodate arbitrary value combinations generated by these configurable factors.
Regarding the MHA design, we allow different configurations of the number of heads (\textit{Attention Head Number}). For the FF block design, we consider two configurable factors: \textit{Intermediate Number}) and \textit{Stacked FF Number}. The former explains the FF intermediate size while the latter illustrates the number of stacked FF networks. Note that every FF network contains two dense layers. 

Possible values of these factors are listed in Table \ref{tab:design_space} as examples 
for quantitative analysis
in the following experiments. In total, this example space contains 576 student model design combinations for one building block. 
Note that {\ourname} supports configurable student model design spaces, which can be further expanded by adding more choices in each configurable factor.


\begin{table}[t]
\vspace{-4pt}
\caption{A five-dimensional student model architecture design space. Note that these only serve as examples and the design space can be configured differently based on user requirements.}
\label{tab:design_space}
\begin{center}
\begin{small}
\begin{tabular}{cll}
\toprule
\# & Configurable Factors & Value Choices \\
\midrule
1   & \textit{Hidden Size} & [128, 246, 384, 512]\\
2   & \textit{Bottleneck Size} & [64, 96, 128, 160] \\
3   & \textit{Attention Head Number} & [1, 2, 4, 8] \\
4   & \textit{Intermediate Number} & [384, 512, 640] \\
5   & \textit{Stacked FF Number} & [2, 4, 6] \\
\bottomrule
\end{tabular}
\end{small}
\end{center}
\vskip -0.15in
\end{table}

\subsection{Flash Distillation}
\label{sec:flash_disllation}

Model accuracy is considered as one of the key metrics to evaluate model candidates. That means the ability to quickly determine a model's accuracy potential is crucial to the search efficiency of an end-to-end framework.
Therefore, we propose Flash Distillation, a model-agnostic knowledge distillation technique that does fast model pre-training to identify promising models with greater potential for high accuracy. Flash Distillation incorporates multiple distillation techniques: we adopt layer-wise knowledge transfer for MHA blocks and building block feature maps by 
similar strategies proposed in \cite{sun2019patient,sanh2019distilbert}; and we use the progressive knowledge transfer proposed by \cite{sun2020mobilebert} to secure an effective knowledge transfer even for deeper models.
 
\subsubsection{Multiple knowledge transfer schemes}
\label{sec:proposeddesign:schemes}
{\ourname} adopts three popular types of knowledge transfer proposed from \cite{sun2020mobilebert}: 
MHA transfer $\mathcal{L}_{MHA} $, building block feature map transfer $\mathcal{L}_{FM} $, and the conventional logit transfer during pre-training distillation $\mathcal{L}_D$. 
Note that {\ourname} is not restricted by the following loss function designs (Equation \ref{eq:att_loss} - \ref{eq:distill_loss}) as they can also be configurable by experienced users.


Since the attention mechanism is the most unique feature in BERT-like models, we enable MHA knowledge transfer to better guide student models in imitating their teacher's attention block behavior. We adopt Kullback-Leibler divergence (KDL), which is a relative entropy, to measure the differences between the student and the teacher ($D_{KDL}$), and our goal is to minimize the loss as shown as follows:
\begin{equation}
\label{eq:att_loss}
    \mathcal{L}_{MHA}^k  = \frac{1}{SH}\sum_{s=1}^{S}\sum_{h=1}^{H} D_{KDL}(a^{\mathcal{T},k}_{s,h} || a^{\mathcal{S},k}_{s,h}),
\end{equation}
where $S$ is the sequence length and $H$ is the number of attention heads. $a^{\mathcal{T}}$ and $a^{\mathcal{S}}$ denote the attention feature maps from the teacher and the student model, respectively. $k$ means the distillation happens in the $k$-th building block.

Regarding the building block feature map transfer, we minimize the mean squared error between the teacher and the student model, which is shown as follows:
\begin{equation}
\label{eq:fm_loss}
    \mathcal{L}_{FM}^k  = \frac{1}{SN}\sum_{s=1}^{S}\sum_{i=1}^{N} (f^{\mathcal{T},k}_{s,i} - f^{\mathcal{S},k}_{s,i})^2,
\end{equation}
where $f^{\mathcal{T}}$ is the teacher's feature map and $f^{\mathcal{S}}$ is the student's feature map. $N$ denotes the feature map size.

After performing layer-wise knowledge transfers, we include model pre-training distillation and use $\mathcal{L}_D$ to represent the distillation loss:
\begin{equation}
\label{eq:distill_loss}
    \mathcal{L}_{D}  = \alpha \mathcal{L}_{M} + (1-\alpha) \mathcal{L}_{MD} + \mathcal{L}_{N},
\end{equation}
where $\mathcal{L}_{M}$ and $\mathcal{L}_{MD}$  denote the Masked Language Modeling (MLM) loss and the MLM distillation loss. $\mathcal{L}_{N}$ is the Next Sentence Prediction (NSP) loss. $\alpha$ is a hyperparameter between 0 and 1.

\subsubsection{Knowledge transfer with different layer dimensions}
Previous work assumes that the teacher and student layers being distilled share the same layer dimension sizes as they are handcrafted with careful considerations of model architectures. However, it is not a reasonable assumption in {\ourname} since we need to target a much broader design space with arbitrary student model architectures. The layer dimension size of the student model may not always be the same as that of the teacher.
To address this issue, we insert a dense layer between the teacher and the student layers where knowledge transfer occurs. 
For example, if the dimension of the student building block output is not the same as the teacher's, a dense layer is inserted between them.
With these additional dense layers, student layers which need to be distilled can be scaled up to match the size of the corresponding teacher layers, so that teacher's knowledge can be transferred smoothly.
We have seen a similar solution adopted by \cite{romero2014fitnets} using an additional convolution layer to match the feature map size.
Layer dimension mismatch also happens between the embedding layer and the first building block because the embedding layer is directly copied from the teacher model. To solve this problem, we down- or up-sample the embedding weights to match the student layer size.

\subsubsection{Progressive knowledge transfer}
\vspace{-2pt}
With the mismatch problem solved, we launch the progressive knowledge transfer to help student models quickly acquire knowledge from the teacher model, which was proposed by \cite{sun2020mobilebert}.
Assuming a student model with $K$ building blocks, layer-wise transfers for MHA and building block feature maps are conducted one block after another. It is a $K$-stage process following the order from block 1 to $K$. When working on the $k$-th stage, all trainable parameters in stage 1 to stage $k-1$ are frozen, so the student model can learn knowledge progressively one building block after another.
Next, we continue the model pre-training distillation until the training reaches the preset training step. In addition, we set a fixed ratio between the step number in progressive distillation and those in the pre-training distillation to guarantee that all three types of distillation schemes mentioned in Section~\ref{sec:proposeddesign:schemes} can be applied effectively.

\subsubsection{Flash distillation vs. regular distillation}
\vspace{-2pt}
In the proposed Flash Distillation, the training step count is set to a much smaller number compared to the regular distillation because they are designed for different goals. Flash distillation works for the early selection of promising student model architectures. It is not necessary to fully train a model to start evaluating it and making decisions. In contrast, regular distillation works for fully preparing student models and make them ready for use. It generally contains hundreds of thousands of steps. We will have more discussions in the experiment (Section \ref{sec:exp_flash_distill}) to illustrate how we select the step number for Flash Distillation.

\subsection{Hardware Performance Integration}
\label{sec:hw_feedback}
\vspace{-2pt}

The basic idea of capturing valid and precise hardware feedback is to measure the student model in the target hardware and software environment in datacenters.
In \textit{Evaluation} stage, the performance measuring process is fully automated and integrated with the rest of the framework.
%
Each model is measured with multiple runs for forward propagation. In each run, the hardware traces are collected, which include inference latency, throughput, accelerator FLOPS utilization, as well as memory capacity and bandwidth utilization.
Particularly, our experiments use the average serving latency as the desired metric, and it can easily be extended to incorporate more metrics in the search engine because the BO algorithm is very flexible for optimizing more objectives.
The collected metrics are then passed to \textit{Model Exploration} for guiding the search process.
Since the hardware performance evaluation is independent of Flash Distillation or pre-training accuracy, it can be completed in parallel to or offline of the distillation process.

\vspace{-2pt}
\subsection{NAS for student models}
\label{sec:nas_algorithm}
\vspace{-2pt}

{\ourname} formulates the student model architecture search as a black-box optimization problem, as it needs minimal assumptions about the problem and minimal internal information of the system~\cite{bergstra2011algorithms, shahriari2015taking}, i.e., the inner relationship between the selected model architecture and its performance objectives.
To be more specific, {\ourname} optimizes the following problem
\begin{equation}
\begin{array}{l}
    \mbox{maximize } {f(x): X \rightarrow \mathbb{R}^o}, \ \
    s.t. \ \
    x \in X
\end{array}
\end{equation}
where $x$ represents a set of configurable factors for describing model architectures and $o$ is the number of objectives. In our experiments, $o$ is two, representing accuracy and latency objectives. \ourname can be easily customized for more and different objectives. With this formulation, all we need is to choose samples (e.g., any $x \in X$) and evaluate $f(x)$, without needing to access other information.

To solve this problem, {\ourname} leverages Bayesian Optimization (BO), an effective black-box optimization algorithm which does not assume any functional forms of objective problems.
It uses Gaussian Processes to learn the posterior distribution of the objective function, which is then used to construct an acquisition function to determine the next trial~\cite{snoek2012practical}.
BO is widely used in 
NAS
~\cite{white2019bananas}, deep learning hyperparameter tuning~\cite{golovin2017vizier,shahriari2015taking}, system optimization~\cite{lagar2019software,dalibard2017boat}, model selection~\cite{malkomes2016bayesian}, transfer learning~\cite{ruder2017learning} and many more~\cite{archetti2019bayesian,srinivas2009gaussian, hutter2011sequential, snoek2012practical, snoek2015scalable, wilson2016deep} for optimizing with limited computing and time budgets.

We use the BO algorithm implemented in Vizier~\cite{golovin2017vizier},
a cloud-based black-box optimization service, and integrates it into the search engine for student architecture search.
Results in Section \ref{sec:exp_search} show that BO outperforms other algorithms that support multi-objective optimization, including random search and evolutionary algorithms.

In previous designs, the gradient-based method is adopted to speedup NAS (referred as differentiable NAS, or DNAS) and it has been demonstrated to work well by generating more accurate models for computer vision tasks~\cite{liu2018darts, wu2019fbnet}.
It is difficult for DNAS to handle NLP tasks,
as it is required to train a huge supernet, containing all possible architecture candidates, on the NLP pre-training tasks, which has been proven to be extremely costly~\cite{xu2021nasbert,gao2021autobert}.
Compared to the DNAS approach, our solution has the following major benefits: 1) {\ourname} does not need to spend enormous effort to train a large supernet 
beforehand on NLP pre-training tasks; 2) it can better scale to handle a much larger design space; and 3) it can be easily extended to new objectives and new models with different architecture configurations.

%% file: sec_experiment.tex
\section{Experiments}
\label{sec:experiment}

In this section, we evaluate {\ourname} and demonstrate its effectiveness for exploring hardware-efficient task-agnostic NLP models.
First, we present the compressed task-agnostic models found by {\ourname} with optimized pre-training accuracy and hardware performance. Since the search is enabled by a BO algorithm, we compare it to random and evolutionary algorithms to demonstrate its better search efficiency. 
Next, we fine-tune these models on a downstream task called SQuAD, and compare them to the state-of-the-art distilled BERT models.
We also provide quantitative analysis to show the importance of using multi-objective search and the effectiveness of Flash Distillation.

\begin{table*}[t]
\caption{The student models found by \ourname. 
We train these models using the same pre-training setup as the baseline (with 740k-step pre-training and the batch size of 2048) and present their pre-training accuracy (Masked lm accuracy) and measured inference latency results. 
All models achieve better accuracy and lower latency compared to the baseline. The number shown in parentheses is the result of comparison to the baseline. \ddag denotes our runs with the open-sourced MobileBERT code.
For reference, MLPerf \cite{mattson2019mlperf} uses Masked lm accuracy $=$ 72 as the target pre-training accuracy of BERT$\rm_{LARGE}$.
}
\label{tab:slected_stu_model}
\vskip 0.15in
\begin{center}
\begin{small}
\begin{tabular}{llll}
\toprule
  & \# Param  & Latency & Masked lm accuracy\\
\midrule
MobileBERT\ddag (baseline) & 25.3 M  & 0.65 ms & 69.2 \\ \hline
Model\_512\_128\_1\_384\_4  &  22.2 M \ \ (87.7\%)  & 0.54 ms (1.21$\times$) & 70.0 (101.3\%)\\
Model\_512\_128\_1\_640\_2  &  20.6 M \ \ (81.4\%)  & \textbf{0.45 ms} (1.44$\times$) & 69.5 (100.4\%)\\
Model\_512\_128\_1\_640\_4  & 28.5 M (112.6\%)  & 0.58 ms (1.12$\times$) & \textbf{71.4} (103.2\%)\\
Model\_512\_128\_2\_640\_2  & 20.6 M \ \ (81.4\%)  & 0.49 ms (1.32$\times$) & 70.0 (101.2\%)\\
Model\_512\_128\_4\_512\_2  & \textbf{19.0 M} \ \ (75.1\%)  & 0.54 ms (1.21$\times$) & 69.7 (100.8\%)\\
Model\_512\_128\_4\_640\_2  & 20.6 M \ \ (81.4\%)  & 0.56 ms (1.16$\times$) & 70.3 (101.6\%)\\
Model\_512\_160\_2\_512\_2  & 22.8 M \ \ (90.1\%)  & 0.59 ms (1.09$\times$) & 70.4 (101.8\%)\\
\bottomrule
\end{tabular}
\end{small}
\end{center}
\vskip -0.1in
\end{table*}

\subsection{Experimental Setup}
\label{sec:exp_setup}
To launch \ourname, we first specify the framework inputs as described in Figure \ref{fig:overall_flow}, which include pre-training tasks, design space, target hardware, and target metrics. 
{\ourname} can also support user-defined constraints to help reduce the search space, such as specifying the minimum acceptable hardware performance.
In this experiment, we do not provide any constraints to limit the model exploration.
To demonstrate {\ourname}'s effectiveness on compressing task-agnostic models, we measure student models' pre-training accuracy~\cite{devlin-etal-2019-bert}, which include Masked Language Modeling (MLM) and Next Sentence Prediction (NSP).
We evaluate {\ourname} on the design space described in Table~\ref{tab:design_space} for building block search and stack 24 of the same building blocks to construct every student model. In this experiment, 
{\tpu} is used as the target hardware to perform model inference. 
Inference latency is measured with batch size $=$1, which can be changed to a representative production serving batch size of the user's target model.
Since our goal is to deliver hardware-efficient models with high accuracy, we pass both hardware and software metrics to {\ourname} as optimization objectives, which are precise model inference latency and model pre-training accuracy.

In each iteration, one student model is selected by the \textit{Model Exploration} stage and passed to the \textit{Flash Distillation} stage for rapid model pre-training. The teacher model 
is called IB-BERT$\rm_{LARGE}$, which is a 24-layer BERT-like model with 293M parameters proposed by \cite{sun2020mobilebert}.
The selected student model follows the pre-training schedule introduced in Section \ref{sec:flash_disllation}, which includes the layer-wise knowledge transfer, the progressive knowledge transfer, and the model pre-training distillation. Each building block is quickly trained by 500 steps for layer-wise knowledge transfer, so the progressive knowledge transfer lasts for 500$\times$24$=$12k steps. Next, the student model is pre-trained for 25k steps with 500 warm-up steps. In total, a Flash Distillation contains 37k steps and it is performed on TPU v3~\footnote{\url{https://cloud.google.com/tpu}} chips with a batch size of 2048 and LAMB optimizer \cite{you2020lamb}.
For pre-training data, we follow the same recipe as BERT using the BooksCorpus \cite{zhu2015aligning} and English Wikipedia.

\subsection{Models that outperform MobileBERT}
MobileBERT~\cite{sun2020mobilebert}, the state-of-the-art task-agnostic design for BERT compression, is selected as our baseline. To ensure a fair comparison, 
we run the open-sourced MobileBERT code~\footnote{\url{https://github.com/google-research/google-research/tree/master/mobilebert}} and use the same hyperparameters for MobileBERT pre-training and our models' regular distillation.
All models are trained from scratch.

After Flash Distillation, {\ourname} finds seven compressed models that exhibit higher pre-training accuracy and lower inference latency on the target hardware than MobileBERT with the same Flash Distillation process. We then run regular distillation for these seven models for a thorough training process. 
It is a much longer training process, which includes 240k-step progressive knowledge transfer (10K steps for each layer) and 500k-step model pre-training with 10k warm-up steps. All models are trained on TPU v3 chips with a batch size of 2048.

After regular distillation, we list their accuracy and inference latency in Table \ref{tab:slected_stu_model}. 
The compressed model candidates are encoded with their five architecture configurable factors described in Table \ref{tab:design_space}. For example, Model\_512\_128\_1\_384\_4 encodes a model architecture with \textit{Hidden Size}$=$512, \textit{Bottleneck Size}$=$128, \textit{Attention Head Number}$=$1, \textit{Intermediate Number}$=$384, and \textit{Stacked FF Number}$=$4.

All seven models outperform the baseline regarding both accuracy and inference latency. Among them, Model\_512\_128\_1\_640\_4 achieves the best accuracy, peaking at 71.4. That is 3.2\% higher than the baseline. Model\_512\_128\_1\_640\_2 maintains a competitive accuracy while achieves 1.44$\times$ speedup on inference latency compared to the baseline, while Model\_512\_128\_4\_512\_2 achieves the smallest model size with 19.0M parameters (75.1\% of the baseline).
We observe that models with more parameters do not always have longer inference latency. For example, Model\_512\_128\_1\_640\_4 has 12.6\% more parameters but it performs 1.12$\times$ faster than the baseline.
Such a counter-intuitive observation could be caused by the larger model's better operational intensity and parallelism, or that its computation pattern is better optimized by the software stack. Similar findings are published in \cite{li2021searching}, where models with more parameters can achieve better latency than those with fewer parameters.
This indicates that optimizing for model size during distillation is not sufficient for better serving latency.
This finding also emphasizes the importance of using precise hardware performance instead of proxy metrics to guide the model search.

\vspace{-2pt}
\subsection{Search Algorithm Comparison}
\vspace{-4pt}
\label{sec:exp_search}

\begin{figure}[t]
    \centering
    \vspace{-2pt}
    \includegraphics[width=0.46\textwidth]{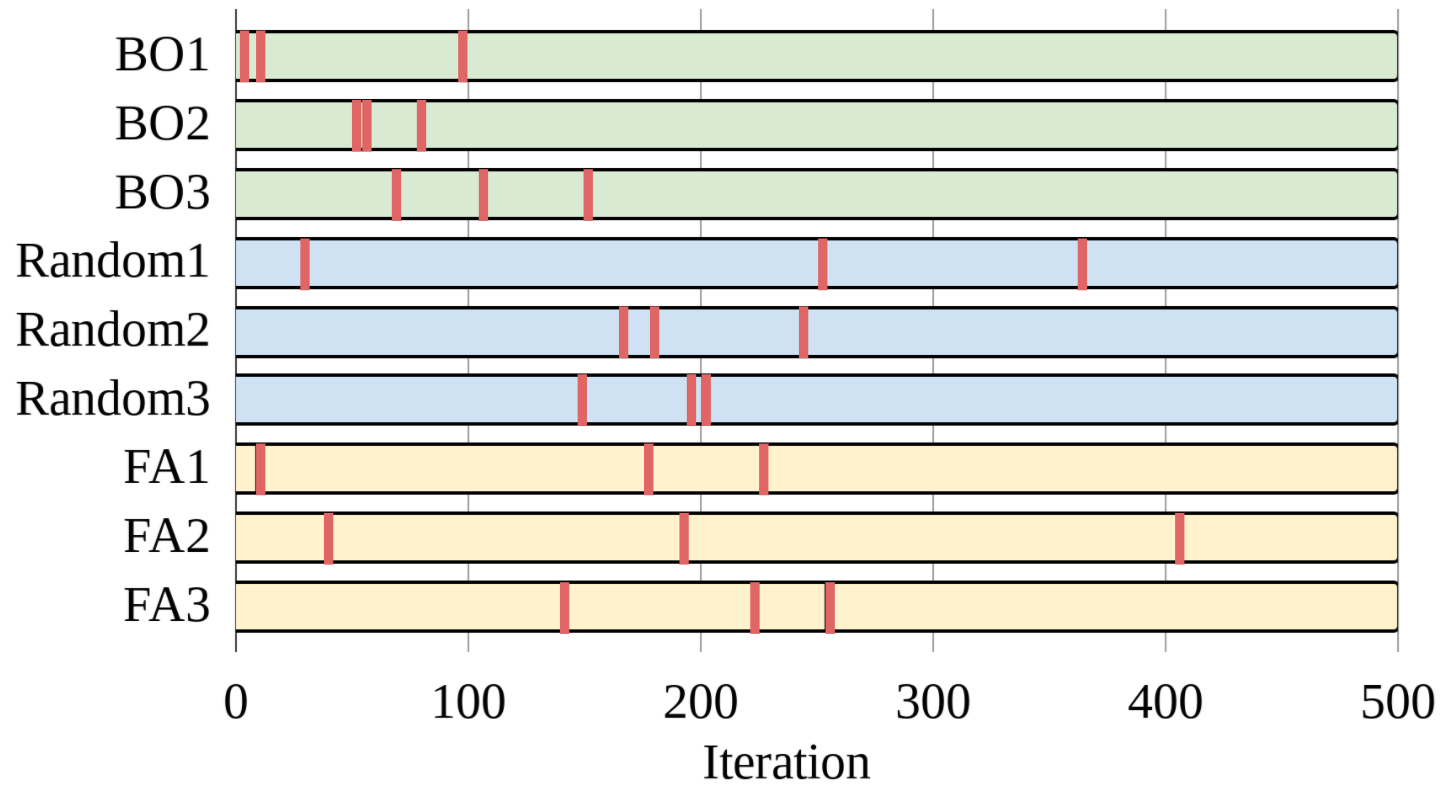}
    \vspace{-8pt}
    \caption{Search efficiency of Bayesian optimization (BO), random search, and firefly algorithm (FA). The three red vertical lines in each bar indicate the iteration numbers where three better-than-MobileBERT student models are found. BO finds three models with fewer iterations than other two algorithms.}
    \label{fig:search_time}
    \vspace{-8pt}
\end{figure}

The search engine in the \textit{Model Exploration} stage is one of the most crucial components facilitating search efficiency of \ourname. 
An efficient search algorithm that gives good candidates with fewer iterations during model search largely reduces the computation cost and end-to-end duration of NAS.
%
To better understand its search efficiency, we compare the BO algorithm, used in \ourname, to other two popular algorithms with the multi-objective support: the random search algorithm and the firefly algorithm (FA) \cite{yang2010firefly}.
The random search represents a naive baseline, which is commonly compared against in previous work~\cite{gao2021autobert}.
In each iteration, random search selects one of the possible models uniformly at random and every trial is independent of other trials.
The FA represents a more sophisticated baseline.
The FA generates a new suggestion every iteration by taking a linear combination of previous suggestions with a small perturbation.

In this experiment, we measure the number of NAS iterations the three search algorithms take to search for three pre-trained models that outperform the baseline model (MobileBERT). For a fair comparison, we maintain the same setup, including the design space, the metrics, and the Flash Distillation technique, while evaluating different search algorithms. 
In \ourname, the search engine proposes one model candidate per iteration, by going through the three major stages of \textit{Model Exploration}, \textit{Flash Distillation}, and \textit{Evaluation}. We set the maximum iteration number as 500 to provide sufficient time for these algorithms and evaluate how many iterations they take to find three models outperforming MobileBERT.

Comparison results are shown in Figure \ref{fig:search_time}, where the x-axis represents the number of iterations and the y-axis indicates nine independent experiments (three trials per algorithm). 
The results show that BO can finish searching much earlier than the other two algorithms. For the best case, it discovers all three models within 80 iterations, and on average, it can finish in 110 iterations. As a comparison, the random algorithm and the FA require 270 and 296 iterations on average to find all three better models.
The average number of iterations to discover the first promising model is another very important metric to compare, because in production, people usually need only one model to deploy.
BO takes 43 iterations on average to find the first promising model, while it is 116 and 64 respectively for random algorithm and FA. 





\subsection{Results on GLUE}
We demonstrate the {\ourname} generated models on the General Language Understanding Evaluation (GLUE) benchmark with nine downstream natural language understanding tasks \cite{wangglue2019}. As shown in Table \ref{tab:glue}, we compare our pre-trained models from Table \ref{tab:slected_stu_model} to BERT$\rm_{BASE}$~\cite{devlin2018bert} and state-of-the-art BERT compression models, including DistillBERT~\cite{sanh2019distilbert}, TinyBERT~\cite{jiao2020tinybert}, NAS-BERT~\cite{xu2021nasbert}, and MobileBERT~\cite{sun2020mobilebert}.

All of the {\ourname} generated models in Table \ref{tab:glue} achieve higher average scores than BERT$\rm_{BASE}$, DistilBERT, TinyBERT$_6$, and MobileBERT with significantly smaller model sizes. Among them, two of our most compact models (Model\_512\_128\_1\_640\_2 and Model\_512\_128\_2\_640\_2) have a 81.1\% size reduction compared to BERT$\rm_{BASE}$ and deliver higher average scores. Our most accurate model (Model\_512\_128\_1\_640\_4) achieves the highest scores in three tasks (CoLA, QQP, and STS-B) and the highest average score (81.69) across all nine tasks. 
It is worth mentioning that {\ourname} is task-agnostic and does not require teacher models (no distillation) for downstream tasks, while TinyBERT and NAS-BERT require teacher models in fine-tuning which have unfair advantages over {\ourname}.

\begin{table*}[t]
\vspace{-8pt}
\caption{The results on the GLUE benchmark. *denotes models conducting knowledge distillation in both pre-training and fine-tuning stages. $\circ$marks MobileBERT without operational optimizations. $\triangleright$marks NAS-BERT with data augmentation.}
\label{tab:glue}
\vspace{-8pt}
\begin{center}
\begin{small}
\begin{tabular}{l|cc|cccccccc|c}
\toprule
  & \# Param  & Latency & CoLA & MNLI-m/mm & MRPC & QNLI & QQP & RTE & SST-2 & STS-B & Avg.  \\
\midrule
BERT$\rm_{BASE}$    & 109 M & - & 52.1 & 84.6/83.4 & \textbf{88.9} & 90.5 & 71.2 & 66.4 & \textbf{93.5} & 85.8 & 79.60 \\ \hline
DistilBERT          & 67 M & -  & 51.3 & 82.2 & 87.5 & 89.2 & 88.5 & 59.9 & 91.3 & 86.9 & 79.60 \\
TinyBERT$\rm_6$*    & 67 M & -  & 51.1 & 84.6/83.2 & 87.3 & 90.4 & 71.6 & 70.0 & 93.1 & 83.7 & 79.44 \\
NAS-BERT*           & 60 M & -  &  48.4 & 83.5 & 84.5 & 90.9 & 88.9 & \textbf{73.7}  & 92.9 & 86.1 & 81.11 \\
NAS-BERT*$\triangleright$       & 60 M & -  & 50.5 & 84.1 & 86.4 & 91.2 & 88.8 & 72.7 & 92.6 & 86.9 & 81.65 \\
MobileBERT          & 25.3 M    & 0.65 ms  & 50.5 & 83.3/82.6 & 88.8 & 90.6 & 70.2 & 66.2 & 92.8 & 84.4 & 78.82\\ 
MobileBERT$\circ$  & 25.3 M    & 0.65 ms  & 51.1 & 84.3/83.4 & 88.8 & \textbf{91.6} & 70.5 & 70.4 & 92.6 & 84.8 & 79.72\\ \hline
Model\_512\_128\_1\_640\_2  & \textbf{20.6 M} & \textbf{0.45 ms} & 53.2 & 81.0/81.9 & 84.1 & 88.9 & 89.4 & 67.2 & 90.8 & 87.0 & 80.38 \\ 	
Model\_512\_128\_2\_640\_2  & \textbf{20.6 M} & 0.49 ms & 53.2 & 82.3/82.5 & 83.8 & 90.2 & 89.7 & 66.1 & 90.8 & 88.1 & 80.75 \\
Model\_512\_160\_2\_512\_2  & 22.8 M & 0.59 ms & 52.1 & 82.6/82.9 & 86.3 & 90.4 & 90.0 & 63.9 & 91.2 & 88.7 & 80.89 \\
Model\_512\_128\_1\_640\_4  & 28.5 M & 0.58 ms &  \textbf{55.9} & 82.7/82.8 & 87.5 & 90.4 & \textbf{90.2} & 66.1 & 90.8 & \textbf{88.8} & \textbf{81.69} \\

\bottomrule
\end{tabular}
\end{small}
\end{center}
\vskip -0.1in
\end{table*}

\subsection{Results on SQuAD}
\begin{table}[t]
\vspace{-8pt}
\caption{The results on SQuAD v1.1. *denotes models conducting knowledge distillation in both pre-training and fine-tuning stages. \dag denotes a model using 1.6$\times$ more pre-training steps than its original setup. \ddag marks our run with the open-sourced MobileBERT code without hyperparameter tuning and using the same fine-tuning setup as our models.}
\label{tab:squad}
\vskip 0.15in
\begin{center}
\begin{small}
\begin{tabular}{lcccc}
\toprule
  & \# Param  & Latency &  F1  & EM\\
\midrule
BERT$\rm_{BASE}$  & 109 M & - & 88.5 & 80.8\\ \hline
DistilBERT          & 67 M & -        & 85.8 & 77.1 \\
DistilBERT*      & 67 M & -        & 86.9 & 79.1 \\
TinyBERT$\rm_6$*            & 67 M & -        & 87.5 & 79.7 \\
NAS-BERT*          & 60 M & -        & 88.0 & 80.5 \\
NAS-BERT*\dag       & 60 M & -        & 88.4 & 81.2 \\ 
MobileBERT          & 25.3 M          & 0.65 ms  & \textbf{90.0} & \textbf{82.9} \\ 
MobileBERT\ddag     & 25.3 M          & 0.65 ms  & 87.7 & 80.0 \\ \hline

Ours-1  & 22.8 M  &  0.59 ms  & 88.4 & 80.8 \\
Ours-2  & \textbf{20.6 M}  &  \textbf{0.49 ms}  & 88.1 & 80.5 \\
\bottomrule
\end{tabular}
\end{small}
\end{center}
\vskip -0.1in
\end{table}

To further evaluate the model quality that {\ourname} found, we evaluate the pre-trained models from Table~\ref{tab:slected_stu_model} with a downstream NLP task called Stanford Question Answering Dataset (SQuAD) \cite{rajpurkar2016squad}. It is a large-scale dataset with 100k crowd-sourced question/answer pairs for question answering and reading comprehension. We choose the dev F1 and the exact match (EM) as accuracy metrics.

Table~\ref{tab:squad} compares our models to 
the BERT$\rm_{BASE}$ \cite{devlin2018bert} and
the recently published compressed designs, including DistillBERT~\cite{sanh2019distilbert}, TinyBERT~\cite{jiao2020tinybert}, NAS-BERT~\cite{xu2021nasbert}, and MobileBERT~\cite{sun2020mobilebert}.
Note that DistillBERT performs knowledge distillation only in model pre-training stage, while DistillBERT* (distinguished with *) uses two-stage knowledge distillation for both model pre-training and fine-tuning. The TinyBERT$\rm_6$* denotes the 6-layer model with two-stage distillation and it is the most accurate model proposed in \cite{jiao2020tinybert}. Similarly, both NAS-BERT designs do distillation in both stages and NAS-BERT*\dag uses 1.6$\times$ more pre-training steps than NAS-BERT*. Such a two-stage distillation strategy generally involves a more complicated distillation pipeline as it needs additional fine-tuned teachers for different downstream tasks. 

In contrast to the above-mentioned designs, MobileBERT and our solutions only require a single-stage knowledge distillation during pre-training, so that the pre-trained models can be directly fine-tuned for downstream tasks. 
We collect MobileBERT performance from \cite{sun2020mobilebert}, which involves hyperparameter tuning especially for the SQuAD task to achieve better accuracy.
In addition to that, we run its open-sourced code following our setups in Section~\ref{sec:exp_setup} without hyperparameter tuning and present the results with the mark \ddag.
The reason why MobileBERT\ddag's results are different from MobileBERT~\cite{sun2020mobilebert} is that we do not conduct hyperparameter tuning as the original paper does to make it a fair comparison for the rest of the models.

Table \ref{tab:squad} shows that the student models found by {\ourname} have lower inference latency, smaller model sizes, while maintaining great accuracy for SQuAD task. 
The average F1 and EM of the seven models are 88 and 80, respectively. Among them, the most accurate one (Ours-1: Model\_512\_160\_2\_512\_2) achieves the same F1 score (88.4) with only 38\% of the parameters compared to NAS-BERT*\dag. The more efficient model (Ours-2: Model\_512\_128\_2\_640\_2) has more compact architecture with only 20.6M parameters and it also has higher accuracy than five other compressed models listed in Table~\ref{tab:squad}.


\subsection{Multi-objective vs. single-objective search}

\begin{figure}[t]
    \centering
    \includegraphics[width=0.45\textwidth]{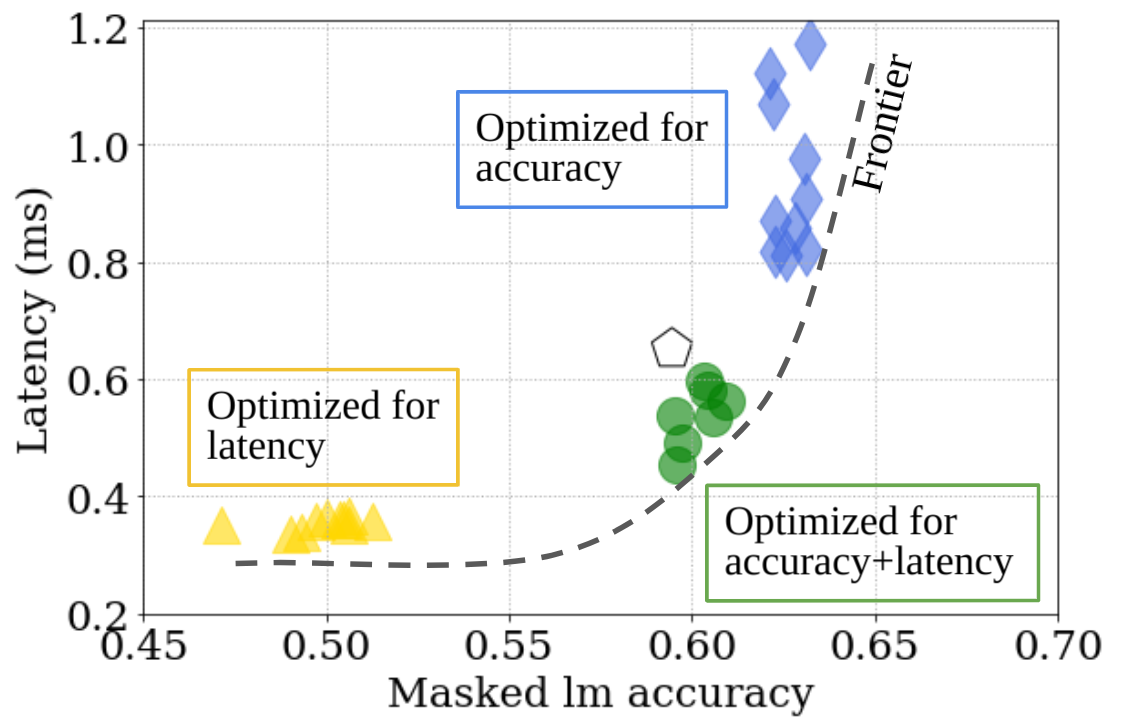}
    \caption{Multi-objective search in {\ourname} helps identify the most promising models by considering both hardware and software metrics after Flash Distillation. The pentagon marks the performance of the MobileBERT representing the baseline. The clusters with yellow triangles and blue diamonds show the models optimized for latency and accuracy, respectively. The green dots at the bottom right are the seven models listed in Table \ref{tab:slected_stu_model}, which are discovered by multi-objective search in \ourname.}
    \label{fig:multiobj_search}
\end{figure}

Additionally, we compare the multi-objective and the single-objective search and show their different effects for compressed model exploration.
The multi-objective search experiment maximizes accuracy and minimizes latency. And the two experiments of single-objective search respectively optimize for model accuracy (which is more popular in previous work) and serving latency. The same model architecture design space in Table \ref{tab:design_space} is used for all three search strategies.

Figure \ref{fig:multiobj_search} shows the most promising models found, with accuracy shown on the x-axis and latency on the y-axis.
We observe totally different behaviors from these search objectives. The single-objective approaches output models that are located close to the two ends of the Pareto curve: one along the lower latency limit (denoted as yellow triangles for optimized latency) and the other along the upper accuracy limit (denoted as blue diamonds for optimized accuracy). 
However, 
these models may not be ideal for production deployment which usually needs to accommodate more than one optimization objective, such as a better balance between accuracy and latency.
By considering multiple objectives, the search process can find all models near the Pareto curve. Assuming users are interested in models with both high accuracy and low latency, they can select models toward the middle of the Pareto curve with all objectives being optimized. In our experiments, those are models (denoted as green dots) near the lower right.
Compared to models that only focus on higher accuracy (the blue ones), these models (the green ones) have 1.8$\times$ speedup on average with a loss 2.5\% accuracy, which could be valuable for production deployment and would not have been found with single-objective search.
It is important for the multi-objective search to find all models along the Pareto curve so that users can have a better idea about the design space and select ones that are of interest by setting different weights for the objectives.

Besides the unconstrained search introduced in the last paragraph, another common single-objective search approach considers metric constraints. For example, maximizing accuracy with a latency constraint, or vise versa.
However, this approach needs reasonable constraints to find models similar to what can be found by multi-objective search and setting constraints is difficult without prior knowledge of the Pareto curve.
Let us use the results in Figure \ref{fig:multiobj_search} as examples and assume a single-objective search tries to maximize accuracy with a latency constraint.
If the latency constraint is lower than 1ms, the search results are similar to optimizing for accuracy only, which are some of the blue diamonds.
If the latency constraint is lower than 0.4ms, we would get models with worse accuracy than the baseline.
It is much easier to adopt multi-objective search to find the student models (in green dots) which are better than the baseline.



\subsection{Flash Distillation}
\label{sec:exp_flash_distill}

\begin{figure}[t]
    \centering
    \includegraphics[width=0.47\textwidth]{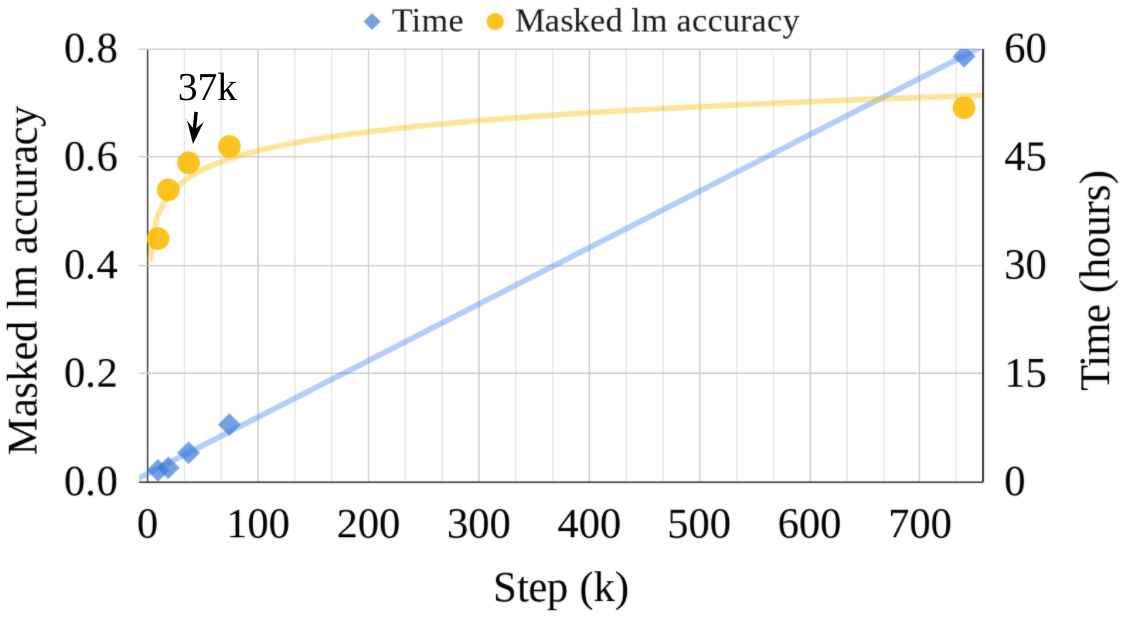}
     \vspace{-4pt}
    \caption{Illustration of four Flash Distillation cases with 9.25k, 18.5k, 37k, and 74k steps comparing to the regular distillation case with 740k steps for the same student model (a total of five data points). 
    The growth of accuracy shows a logarithmic trend while the time spent on training increases linearly with increasing training steps.}
    \vspace{-8pt}
    \label{fig:time_vs_accuracy}
\end{figure}

\begin{figure}[t]
    \centering
     \vspace{-8pt}
    \includegraphics[width=0.44\textwidth]{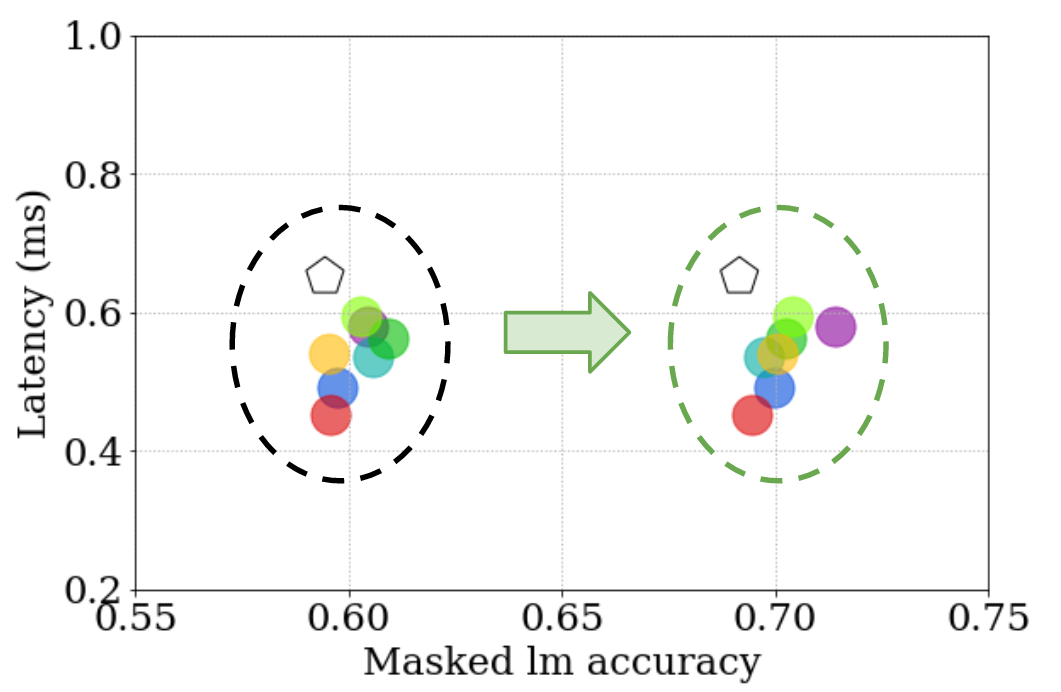}
    \vspace{-4pt}
    \caption{The models selected using Flash Distillation with 37k steps (the left cluster) and the same ones using regular distillation with 740k steps (the right cluster). Promising models can be identified much earlier using Flash Distillation.
    Note that the pentagon denotes the performance of MobileBERT (baseline).
    }
    \vspace{-8pt}
    \label{fig:flash_distill}
\end{figure}

With Flash Distillation, {\ourname} is able to select models with large potential to achieve high accuracy at an early pre-training stage.
We evaluate the accuracy potential of student models with a smaller number of training steps during knowledge transfer, and skip unpromising candidates for regular distillation.

To determine a suitable step number, we launch four Flash Distillation tests with 1.25\%, 2.5\%, 5\%, and 10\% of the step number in the regular distillation (which contains 740k steps). The ratio between progressive and pre-training distillation is set to 0.48 for all cases. We plot the pre-training accuracy and distillation time for each case in Figure \ref{fig:time_vs_accuracy}. Since the accuracy shows a logarithmic trend, selecting 
a spot when the accuracy growth becomes slower seems a great trade-off between training costs and accuracy gain. We therefore select 37k (5\% of the regular distillation steps) as the Flash Distillation step number.

To verify that Flash Distillation is effective in selecting promising candidates, we perform flash and regular distillation for models from Table~\ref{tab:slected_stu_model} and compare the changes in their relative positions in accuracy and latency dimensions. 
Figure \ref{fig:flash_distill} shows the results of the same group of models being pre-trained for 37k (in Flash Distillation) and 740k steps (in regular distillation), respectively. Each dot represents one student model in Table \ref{tab:slected_stu_model} and the pentagon denotes the baseline model (MobileBERT). Since different distillation methods for the same model only affect model accuracy, we observe that dots are shifted to the right, meaning they become more accurate after a much longer pre-training process. Their relative positions, however, are fairly stable regardless of using Flash Distillation or regular distillation. 
In this experiment, Flash Distillation only consumes 5\% of the training steps to grow the models compared to using regular distillation and models can already be distinguished even though have not reached the final accuracy.
We therefore conclude that the Flash Distillation is helpful to facilitate the early selection of promising models.



%% file: sec_conclusion.tex
\vspace{-2pt}
\section{Conclusion}
\label{sec:conclusion}
\vspace{-2pt}
This paper presented {\ourname}, an end-to-end model distillation framework that integrates model architecture exploration and multi-objective optimization to explore hardware-efficient task-agnostic NLP models for datacenter deployment.
We formulated the compressed model exploration as a black-box optimization problem and adopted BO to conduct multi-objective model architecture search. To rapidly identify more promising models, 
we introduced a model-agnostic pre-training technique, called Flash Distillation, to enable fast knowledge distillation for compressed model candidates. 
Experiments on {\tpu} showed that models generated by {\ourname} achieved up to 3.2\% higher pre-trained accuracy and up to 1.44$\times$ speedup on latency compared MobileBERT~\cite{sun2020mobilebert}. 
By evaluating on GLUE, our most accurate model with 28.5M parameters achieved an 81.69 average score, which outperformed BERT$\rm_{BASE}$~\cite{devlin2018bert}, DistillBERT~\cite{sanh2019distilbert}, TinyBERT~\cite{jiao2020tinybert}, NAS-BERT~\cite{xu2021nasbert}, and MobileBERT. Our most compact model with 20.6M parameters (81.1\% size reduction compared to BERT$\rm_{BASE}$) still achieved a higher average score than BERT$\rm_{BASE}$, DistillBERT, TinyBERT, and MobileBERT.
By evaluating on SQuAD, two of the proposed models with less than 23M parameters achieved higher accuracy than DistillBERT, TinyBERT, and NAS-BERT. 

%% file: sec_acknowledge.tex
\section{Acknowledgements}
We would like to thank
Kaya Bekiroğlu,
Maarten Bosma,
Chiachen Chou,
Yanping Huang,
Naveen Kumar,
Yaguang Li,
CK Luk,
Peter Ma,
Martin Maas,
Peter Mattson,
Mangpo Phothilimthana,
David So,
Kefan Xiao, and
Hongkun Yu
for their feedback.

%% file: example_paper.bbl
\begin{thebibliography}{48}
\providecommand{\natexlab}[1]{#1}
\providecommand{\url}[1]{\texttt{#1}}
\expandafter\ifx\csname urlstyle\endcsname\relax
  \providecommand{\doi}[1]{doi: #1}\else
  \providecommand{\doi}{doi: \begingroup \urlstyle{rm}\Url}\fi

\bibitem[Archetti \& Candelieri(2019)Archetti and
  Candelieri]{archetti2019bayesian}
Archetti, F. and Candelieri, A.
\newblock \emph{Bayesian optimization and data science}.
\newblock Springer, 2019.

\bibitem[Bergstra et~al.(2011)Bergstra, Bardenet, Bengio, and
  K{\'e}gl]{bergstra2011algorithms}
Bergstra, J., Bardenet, R., Bengio, Y., and K{\'e}gl, B.
\newblock Algorithms for hyper-parameter optimization.
\newblock \emph{Advances in neural information processing systems}, 2011.

\bibitem[Chen et~al.(2020)Chen, Li, Qiu, Wang, Li, Ding, Deng, Huang, Lin, and
  Zhou]{chen2020adabert}
Chen, D., Li, Y., Qiu, M., Wang, Z., Li, B., Ding, B., Deng, H., Huang, J.,
  Lin, W., and Zhou, J.
\newblock {AdaBERT}: Task-adaptive {BERT} compression with differentiable
  neural architecture search.
\newblock In \emph{Proceedings of the Twenty-Ninth International Joint
  Conference on Artificial Intelligence}, pp.\  2463--2469. International Joint
  Conferences on Artificial Intelligence Organization, 2020.
\newblock \doi{10.24963/ijcai.2020/341}.
\newblock URL \url{https://doi.org/10.24963/ijcai.2020/341}.

\bibitem[Dalibard et~al.(2017)Dalibard, Schaarschmidt, and
  Yoneki]{dalibard2017boat}
Dalibard, V., Schaarschmidt, M., and Yoneki, E.
\newblock Boat: Building auto-tuners with structured bayesian optimization.
\newblock In \emph{Proceedings of the 26th International Conference on World
  Wide Web}, pp.\  479--488, 2017.

\bibitem[Devlin et~al.(2018)Devlin, Chang, Lee, and Toutanova]{devlin2018bert}
Devlin, J., Chang, M.-W., Lee, K., and Toutanova, K.
\newblock Bert: Pre-training of deep bidirectional transformers for language
  understanding.
\newblock \emph{arXiv preprint arXiv:1810.04805}, 2018.

\bibitem[Devlin et~al.(2019)Devlin, Chang, Lee, and
  Toutanova]{devlin-etal-2019-bert}
Devlin, J., Chang, M.-W., Lee, K., and Toutanova, K.
\newblock {BERT}: Pre-training of deep bidirectional transformers for language
  understanding.
\newblock In \emph{Proceedings of the Conference of the North {A}merican
  Chapter of the Association for Computational Linguistics: Human Language
  Technologies}, pp.\  4171--4186, Minneapolis, Minnesota, 2019. Association
  for Computational Linguistics.
\newblock \doi{10.18653/v1/N19-1423}.
\newblock URL \url{https://aclanthology.org/N19-1423}.

\bibitem[Floridi \& Chiriatti(2020)Floridi and Chiriatti]{floridi2020gpt}
Floridi, L. and Chiriatti, M.
\newblock Gpt-3: Its nature, scope, limits, and consequences.
\newblock \emph{Minds and Machines}, 30\penalty0 (4):\penalty0 681--694, 2020.

\bibitem[Gao et~al.(2021)Gao, Xu, Ren, Yu, Liang, Jiang, Li,
  et~al.]{gao2021autobert}
Gao, J., Xu, H., Ren, X., Yu, P.~L., Liang, X., Jiang, X., Li, Z., et~al.
\newblock {AutoBERT-Zero}: Evolving bert backbone from scratch.
\newblock \emph{arXiv preprint arXiv:2107.07445}, 2021.

\bibitem[Golovin et~al.(2017)Golovin, Solnik, Moitra, Kochanski, Karro, and
  Sculley]{golovin2017vizier}
Golovin, D., Solnik, B., Moitra, S., Kochanski, G., Karro, J., and Sculley, D.
\newblock Google vizier: A service for black-box optimization.
\newblock In \emph{Proceedings of the 23rd ACM SIGKDD international conference
  on knowledge discovery and data mining}, pp.\  1487--1495, 2017.

\bibitem[Gordon et~al.(2020)Gordon, Duh, and Andrews]{gordon2020compressing}
Gordon, M., Duh, K., and Andrews, N.
\newblock Compressing {BERT}: Studying the effects of weight pruning on
  transfer learning.
\newblock In \emph{Proceedings of the 5th Workshop on Representation Learning
  for NLP}, pp.\  143--155, 2020.

\bibitem[Hinton et~al.(2015)Hinton, Vinyals, and Dean]{hinton2015distilling}
Hinton, G., Vinyals, O., and Dean, J.
\newblock Distilling the knowledge in a neural network.
\newblock \emph{arXiv preprint arXiv:1503.02531}, 2015.

\bibitem[Hou et~al.(2020)Hou, Huang, Shang, Jiang, Chen, and
  Liu]{hou2020dynabert}
Hou, L., Huang, Z., Shang, L., Jiang, X., Chen, X., and Liu, Q.
\newblock {DynaBERT}: Dynamic bert with adaptive width and depth.
\newblock \emph{Advances in Neural Information Processing Systems}, 2020.

\bibitem[Hutter et~al.(2011)Hutter, Hoos, and
  Leyton-Brown]{hutter2011sequential}
Hutter, F., Hoos, H.~H., and Leyton-Brown, K.
\newblock Sequential model-based optimization for general algorithm
  configuration.
\newblock In \emph{International conference on learning and intelligent
  optimization}, pp.\  507--523. Springer, 2011.

\bibitem[Jiao et~al.(2020)Jiao, Yin, Shang, Jiang, Chen, Li, Wang, and
  Liu]{jiao2020tinybert}
Jiao, X., Yin, Y., Shang, L., Jiang, X., Chen, X., Li, L., Wang, F., and Liu,
  Q.
\newblock {TinyBERT}: Distilling bert for natural language understanding.
\newblock In \emph{Proceedings of the 2020 Conference on Empirical Methods in
  Natural Language Processing: Findings}, pp.\  4163--4174, 2020.

\bibitem[Jouppi et~al.(2021)Jouppi, Yoon, Ashcraft, Gottscho, Jablin, Kurian,
  Laudon, Li, Ma, Ma, et~al.]{jouppi2021ten}
Jouppi, N.~P., Yoon, D.~H., Ashcraft, M., Gottscho, M., Jablin, T.~B., Kurian,
  G., Laudon, J., Li, S., Ma, P., Ma, X., et~al.
\newblock Ten lessons from three generations shaped google’s tpuv4i:
  Industrial product.
\newblock In \emph{Proceedings of the Annual International Symposium on
  Computer Architecture (ISCA)}, pp.\  1--14, 2021.

\bibitem[Lagar-Cavilla et~al.(2019)Lagar-Cavilla, Ahn, Souhlal, Agarwal, Burny,
  Butt, Chang, Chaugule, Deng, Shahid, et~al.]{lagar2019software}
Lagar-Cavilla, A., Ahn, J., Souhlal, S., Agarwal, N., Burny, R., Butt, S.,
  Chang, J., Chaugule, A., Deng, N., Shahid, J., et~al.
\newblock Software-defined far memory in warehouse-scale computers.
\newblock In \emph{Proceedings of the Twenty-Fourth International Conference on
  Architectural Support for Programming Languages and Operating Systems}, pp.\
  317--330, 2019.

\bibitem[Lan et~al.(2020)Lan, Chen, Goodman, Gimpel, Sharma, and
  Soricut]{Lan2020albert}
Lan, Z., Chen, M., Goodman, S., Gimpel, K., Sharma, P., and Soricut, R.
\newblock {ALBERT:} {A} lite {BERT} for self-supervised learning of language
  representations.
\newblock In \emph{Proceedings of the International Conference on Learning
  Representations}, 2020.

\bibitem[Li et~al.(2021)Li, Tan, Pang, Li, Cheng, Le, and
  Jouppi]{li2021searching}
Li, S., Tan, M., Pang, R., Li, A., Cheng, L., Le, Q.~V., and Jouppi, N.~P.
\newblock Searching for fast model families on datacenter accelerators.
\newblock In \emph{Proceedings of the IEEE/CVF Conference on Computer Vision
  and Pattern Recognition}, pp.\  8085--8095, 2021.

\bibitem[Liu et~al.(2019{\natexlab{a}})Liu, Simonyan, and Yang]{liu2018darts}
Liu, H., Simonyan, K., and Yang, Y.
\newblock {DARTS}: Differentiable architecture search.
\newblock In \emph{Proceedings of the International Conference on Learning
  Representations}, 2019{\natexlab{a}}.

\bibitem[Liu et~al.(2019{\natexlab{b}})Liu, Ott, Goyal, Du, Joshi, Chen, Levy,
  Lewis, Zettlemoyer, and Stoyanov]{liu2019roberta}
Liu, Y., Ott, M., Goyal, N., Du, J., Joshi, M., Chen, D., Levy, O., Lewis, M.,
  Zettlemoyer, L., and Stoyanov, V.
\newblock Roberta: A robustly optimized bert pretraining approach.
\newblock \emph{arXiv preprint arXiv:1907.11692}, 2019{\natexlab{b}}.

\bibitem[Malkomes et~al.(2016)Malkomes, Schaff, and
  Garnett]{malkomes2016bayesian}
Malkomes, G., Schaff, C., and Garnett, R.
\newblock Bayesian optimization for automated model selection.
\newblock In \emph{Workshop on Automatic Machine Learning}, pp.\  41--47. PMLR,
  2016.

\bibitem[Mattson et~al.(2019)Mattson, Cheng, Coleman, Diamos, Micikevicius,
  Patterson, Tang, Wei, Bailis, Bittorf, Brooks, Chen, Dutta, Gupta, Hazelwood,
  Hock, Huang, Ike, Jia, Kang, Kanter, Kumar, Liao, Ma, Narayanan, Oguntebi,
  Pekhimenko, Pentecost, Reddi, Robie, John, Tabaru, Wu, Xu, Yamazaki, Young,
  and Zaharia]{mattson2019mlperf}
Mattson, P., Cheng, C., Coleman, C., Diamos, G., Micikevicius, P., Patterson,
  D., Tang, H., Wei, G.-Y., Bailis, P., Bittorf, V., Brooks, D., Chen, D.,
  Dutta, D., Gupta, U., Hazelwood, K., Hock, A., Huang, X., Ike, A., Jia, B.,
  Kang, D., Kanter, D., Kumar, N., Liao, J., Ma, G., Narayanan, D., Oguntebi,
  T., Pekhimenko, G., Pentecost, L., Reddi, V.~J., Robie, T., John, T.~S.,
  Tabaru, T., Wu, C.-J., Xu, L., Yamazaki, M., Young, C., and Zaharia, M.
\newblock Mlperf training benchmark, 2019.

\bibitem[Peters et~al.(2018)Peters, Neumann, Iyyer, Gardner, Clark, Lee, and
  Zettlemoyer]{peters2018deep}
Peters, M.~E., Neumann, M., Iyyer, M., Gardner, M., Clark, C., Lee, K., and
  Zettlemoyer, L.
\newblock Deep contextualized word representations.
\newblock In \emph{Proceedings of the Conference of the North {A}merican
  Chapter of the Association for Computational Linguistics: Human Language
  Technologies}, pp.\  2227--2237, 2018.

\bibitem[Rajpurkar et~al.(2016)Rajpurkar, Zhang, Lopyrev, and
  Liang]{rajpurkar2016squad}
Rajpurkar, P., Zhang, J., Lopyrev, K., and Liang, P.
\newblock {SQuAD}: 100,000+ questions for machine comprehension of text.
\newblock In \emph{Proceedings of the 2016 Conference on Empirical Methods in
  Natural Language Processing}, pp.\  2383--2392, 2016.

\bibitem[Romero et~al.(2014)Romero, Ballas, Kahou, Chassang, Gatta, and
  Bengio]{romero2014fitnets}
Romero, A., Ballas, N., Kahou, S.~E., Chassang, A., Gatta, C., and Bengio, Y.
\newblock Fitnets: Hints for thin deep nets.
\newblock \emph{arXiv preprint arXiv:1412.6550}, 2014.

\bibitem[Ruder \& Plank(2017)Ruder and Plank]{ruder2017learning}
Ruder, S. and Plank, B.
\newblock Learning to select data for transfer learning with bayesian
  optimization.
\newblock \emph{arXiv preprint arXiv:1707.05246}, 2017.

\bibitem[Sanh et~al.(2019)Sanh, Debut, Chaumond, and Wolf]{sanh2019distilbert}
Sanh, V., Debut, L., Chaumond, J., and Wolf, T.
\newblock {DistilBERT}, a distilled version of {BERT}: smaller, faster, cheaper
  and lighter.
\newblock \emph{arXiv preprint arXiv:1910.01108}, 2019.

\bibitem[Shahriari et~al.(2015)Shahriari, Swersky, Wang, Adams, and
  De~Freitas]{shahriari2015taking}
Shahriari, B., Swersky, K., Wang, Z., Adams, R.~P., and De~Freitas, N.
\newblock Taking the human out of the loop: A review of bayesian optimization.
\newblock \emph{Proceedings of the IEEE}, 104\penalty0 (1):\penalty0 148--175,
  2015.

\bibitem[Shen et~al.(2020)Shen, Dong, Ye, Ma, Yao, Gholami, Mahoney, and
  Keutzer]{shen2020qbert}
Shen, S., Dong, Z., Ye, J., Ma, L., Yao, Z., Gholami, A., Mahoney, M.~W., and
  Keutzer, K.
\newblock {Q-BERT}: Hessian based ultra low precision quantization of {BERT}.
\newblock In \emph{Proceedings of the AAAI Conference on Artificial
  Intelligence}, pp.\  8815--8821, 2020.

\bibitem[Snoek et~al.(2012)Snoek, Larochelle, and Adams]{snoek2012practical}
Snoek, J., Larochelle, H., and Adams, R.~P.
\newblock Practical bayesian optimization of machine learning algorithms.
\newblock \emph{Advances in neural information processing systems}, 25, 2012.

\bibitem[Snoek et~al.(2015)Snoek, Rippel, Swersky, Kiros, Satish, Sundaram,
  Patwary, Prabhat, and Adams]{snoek2015scalable}
Snoek, J., Rippel, O., Swersky, K., Kiros, R., Satish, N., Sundaram, N.,
  Patwary, M., Prabhat, M., and Adams, R.
\newblock Scalable bayesian optimization using deep neural networks.
\newblock In \emph{International conference on machine learning}, pp.\
  2171--2180, 2015.

\bibitem[Srinivas et~al.(2009)Srinivas, Krause, Kakade, and
  Seeger]{srinivas2009gaussian}
Srinivas, N., Krause, A., Kakade, S.~M., and Seeger, M.
\newblock Gaussian process optimization in the bandit setting: No regret and
  experimental design.
\newblock \emph{arXiv preprint arXiv:0912.3995}, 2009.

\bibitem[Sun et~al.(2019)Sun, Cheng, Gan, and Liu]{sun2019patient}
Sun, S., Cheng, Y., Gan, Z., and Liu, J.
\newblock Patient knowledge distillation for {BERT} model compression.
\newblock In \emph{Proceedings of the 2019 Conference on Empirical Methods in
  Natural Language Processing and the 9th International Joint Conference on
  Natural Language Processing}, pp.\  4323--4332, 2019.

\bibitem[Sun et~al.(2020)Sun, Yu, Song, Liu, Yang, and Zhou]{sun2020mobilebert}
Sun, Z., Yu, H., Song, X., Liu, R., Yang, Y., and Zhou, D.
\newblock {MobileBERT}: a compact task-agnostic {BERT} for resource-limited
  devices.
\newblock In \emph{Proceedings of the 58th Annual Meeting of the Association
  for Computational Linguistics}, pp.\  2158--2170, 2020.

\bibitem[Tang et~al.(2019)Tang, Lu, Liu, Mou, Vechtomova, and
  Lin]{tang2019distilling}
Tang, R., Lu, Y., Liu, L., Mou, L., Vechtomova, O., and Lin, J.
\newblock Distilling task-specific knowledge from {BERT} into simple neural
  networks.
\newblock \emph{arXiv preprint arXiv:1903.12136}, 2019.

\bibitem[Tsai et~al.(2019)Tsai, Riesa, Johnson, Arivazhagan, Li, and
  Archer]{tsai2019small}
Tsai, H., Riesa, J., Johnson, M., Arivazhagan, N., Li, X., and Archer, A.
\newblock Small and practical {BERT} models for sequence labeling.
\newblock In \emph{Proceedings of the 2019 Conference on Empirical Methods in
  Natural Language Processing and the 9th International Joint Conference on
  Natural Language Processing}, pp.\  3632--3636, 2019.

\bibitem[Tsai et~al.(2020)Tsai, Ooi, Ferng, Chung, and Riesa]{tsai2020finding}
Tsai, H., Ooi, J., Ferng, C.-S., Chung, H.~W., and Riesa, J.
\newblock Finding fast transformers: One-shot neural architecture search by
  component composition.
\newblock \emph{arXiv preprint arXiv:2008.06808}, 2020.

\bibitem[Vaswani et~al.(2017)Vaswani, Shazeer, Parmar, Uszkoreit, Jones, Gomez,
  Kaiser, and Polosukhin]{vaswani2017attention}
Vaswani, A., Shazeer, N., Parmar, N., Uszkoreit, J., Jones, L., Gomez, A.~N.,
  Kaiser, {\L}., and Polosukhin, I.
\newblock Attention is all you need.
\newblock In \emph{Advances in neural information processing systems}, pp.\
  5998--6008, 2017.

\bibitem[Wang et~al.(2019)Wang, Singh, Michael, Hill, Levy, and
  Bowman]{wangglue2019}
Wang, A., Singh, A., Michael, J., Hill, F., Levy, O., and Bowman, S.~R.
\newblock {GLUE}: A multi-task benchmark and analysis platform for natural
  language understanding.
\newblock In \emph{Proceedings of the International Conference on Learning
  Representations (ICLR)}, 2019.

\bibitem[White et~al.(2019)White, Neiswanger, and Savani]{white2019bananas}
White, C., Neiswanger, W., and Savani, Y.
\newblock Bananas: Bayesian optimization with neural architectures for neural
  architecture search.
\newblock \emph{arXiv preprint arXiv:1910.11858}, 1\penalty0 (2), 2019.

\bibitem[Wilson et~al.(2016)Wilson, Hu, Salakhutdinov, and
  Xing]{wilson2016deep}
Wilson, A.~G., Hu, Z., Salakhutdinov, R., and Xing, E.~P.
\newblock Deep kernel learning.
\newblock In \emph{Artificial intelligence and statistics}, pp.\  370--378.
  PMLR, 2016.

\bibitem[Wu et~al.(2019)Wu, Dai, Zhang, Wang, Sun, Wu, Tian, Vajda, Jia, and
  Keutzer]{wu2019fbnet}
Wu, B., Dai, X., Zhang, P., Wang, Y., Sun, F., Wu, Y., Tian, Y., Vajda, P.,
  Jia, Y., and Keutzer, K.
\newblock {FBNet}: Hardware-aware efficient {ConvNet} design via differentiable
  neural architecture search.
\newblock In \emph{Proceedings of the IEEE/CVF Conference on Computer Vision
  and Pattern Recognition}, pp.\  10734--10742, 2019.

\bibitem[Xu et~al.(2021)Xu, Tan, Luo, Song, Li, Qin, and Liu]{xu2021nasbert}
Xu, J., Tan, X., Luo, R., Song, K., Li, J., Qin, T., and Liu, T.-Y.
\newblock {NAS-BERT}: Task-agnostic and adaptive-size bert compression with
  neural architecture search.
\newblock In \emph{Proceedings of the 27th ACM SIGKDD Conference on Knowledge
  Discovery \& Data Mining}, pp.\  1933–1943, 2021.
\newblock \doi{10.1145/3447548.3467262}.
\newblock URL \url{https://doi.org/10.1145/3447548.3467262}.

\bibitem[Yang(2010)]{yang2010firefly}
Yang, X.-S.
\newblock Firefly algorithm, stochastic test functions and design optimisation.
\newblock \emph{International journal of bio-inspired computation}, 2\penalty0
  (2):\penalty0 78--84, 2010.

\bibitem[Yang et~al.(2019)Yang, Dai, Yang, Carbonell, Salakhutdinov, and
  Le]{yang2019xlnet}
Yang, Z., Dai, Z., Yang, Y., Carbonell, J., Salakhutdinov, R.~R., and Le, Q.~V.
\newblock {XLNet}: Generalized autoregressive pretraining for language
  understanding.
\newblock \emph{Advances in neural information processing systems}, pp.\
  5753--5763, 2019.

\bibitem[You et~al.(2020)You, Li, Reddi, Hseu, Kumar, Bhojanapalli, Song,
  Demmel, Keutzer, and Hsieh]{you2020lamb}
You, Y., Li, J., Reddi, S., Hseu, J., Kumar, S., Bhojanapalli, S., Song, X.,
  Demmel, J., Keutzer, K., and Hsieh, C.-J.
\newblock Large batch optimization for deep learning: Training {BERT} in 76
  minutes.
\newblock In \emph{Proceedings of the International Conference on Learning
  Representations}, 2020.

\bibitem[Zhang et~al.(2019)Zhang, Han, Liu, Jiang, Sun, and
  Liu]{zhang2019ernie}
Zhang, Z., Han, X., Liu, Z., Jiang, X., Sun, M., and Liu, Q.
\newblock Ernie: Enhanced language representation with informative entities.
\newblock In \emph{Proceedings of the 57th Annual Meeting of the Association
  for Computational Linguistics}, pp.\  1441--1451, 2019.

\bibitem[Zhu et~al.(2015)Zhu, Kiros, Zemel, Salakhutdinov, Urtasun, Torralba,
  and Fidler]{zhu2015aligning}
Zhu, Y., Kiros, R., Zemel, R., Salakhutdinov, R., Urtasun, R., Torralba, A.,
  and Fidler, S.
\newblock Aligning books and movies: Towards story-like visual explanations by
  watching movies and reading books.
\newblock In \emph{Proceedings of the IEEE international conference on computer
  vision}, pp.\  19--27, 2015.

\end{thebibliography}
